\documentclass{article}

\usepackage{times}
\usepackage{graphicx} 
\usepackage{subfigure} 

\usepackage{natbib}

\usepackage{algorithm}
\usepackage{algorithmic}

\usepackage{amsmath}
\usepackage{amssymb}

\usepackage{amsfonts}
\newtheorem{theorem}{Theorem}

\usepackage{hyperref}

\usepackage[accepted]{icml2017}

\icmltitlerunning{Deep Tensor Convolution on Multicores}

\begin{document} 

\twocolumn[
\icmltitle{Deep Tensor Convolution on Multicores}



\icmlsetsymbol{equal}{*}

\begin{icmlauthorlist}
\icmlauthor{David Budden}{mit}
\icmlauthor{Alexander Matveev}{mit}
\icmlauthor{Shibani Santurkar}{mit}
\icmlauthor{Shraman Ray Chaudhuri}{mit}
\icmlauthor{Nir Shavit}{mit}
\end{icmlauthorlist}

\icmlaffiliation{mit}{Massachusetts Institute of Technology}

\icmlcorrespondingauthor{David Budden}{budden@csail.mit.edu}

\icmlkeywords{boring formatting information, machine learning, ICML}

\vskip 0.3in
]



\printAffiliationsAndNotice{}  

\begin{abstract} 
Deep convolutional neural networks (ConvNets) of 3-dimensional kernels allow joint modeling of spatiotemporal features. These networks have improved performance of video and volumetric image analysis, but have been limited in size due to the low memory ceiling of GPU hardware. Existing CPU implementations overcome this constraint but are impractically slow. Here we extend and optimize the faster Winograd-class of  convolutional algorithms to the $N$-dimensional case and specifically for CPU hardware. First, we remove the need to manually hand-craft algorithms by exploiting the relaxed constraints and cheap sparse access of CPU memory. Second, we maximize CPU utilization and multicore scalability by transforming data matrices to be cache-aware, integer multiples of AVX vector widths. Treating 2D ConvNets as a special case, we demonstrate a 5 to 25-fold improvement in throughput compared to previous state-of-the-art. \vspace{-0.7cm}
\end{abstract} 

\section{Introduction}
Although convolutional neural networks (ConvNets) have been successfully applied to solve non-trivial image processing problems since the 1990s~\cite{lecun1989backpropagation,lecun2012efficient}, their adoption as a de facto standard for image classification~\cite{russakovsky2015imagenet} and segmentation~\cite{long2015fully} is due largely to recent breakthroughs in network architecture.  Beginning with AlexNet in 2012~\cite{krizhevsky2012imagenet}, the annual ImageNet classification challenge (ILSVRC) has been dominated by progressively deeper networks with smaller kernels~\cite{szegedy2015going,simonyan2014very}. Recent solutions to the issues of vanishing and exploding gradients~\cite{glorot2010understanding} have allowed these networks to extend even deeper, with the ILSVRC15 winner (ResNet~\cite{he2016deep}) being 8-fold deeper than VGG.

It is easy to see why the ``deeper is better" trend has led to better performing ConvNets. Constructing even a modest 7 x 7 receptive field with stacked $k=3$ kernels requires $45\%$ fewer parameters than a single kernel of size $k=7$. Intuitively it also captures a richer set of features due to additional non-linearity. Recent studies have begun to formalize the expressive power of deep versus shallow networks, finding that classification boundaries acquire local curvature and expressivity as an exponential function of network depth but not breadth~\cite{poole2016exponential}. The only obvious trade-off to this performance is the extra memory necessary to store the intermediate activations in deeper networks.

Motivated by the success of these models in image processing tasks, researchers have begun to investigate ConvNet applications in the video processing domain. Example applications include video classification~\cite{karpathy2014large}, segmentation~\cite{couprie2013indoor} and de-noising~\cite{shi2016real}. An important observation that has emerged from these studies is the importance of 3D convolutional primitives for modelling joint spatio-temporal features; the na\"{\i}ve application of traditional 2D ConvNets frame-by-frame does not capture motion continuity or other rich temporal correlations~\cite{ledig2016photo,tran2015learning}. It is thus unsurprising that simple 3D ConvNets have yielded state-of-the-art performance on video classification benchmarks~\cite{tran2015learning} and volumetric image segmentation, e.g. tracing neurons between electron microscopy samples~\cite{lee2015recursive}.

Given the early success and conceptual simplicity of 3D ConvNets, it is interesting to note that many popular deep learning libraries (e.g. Caffe~\cite{jia2014caffe}) do not provide native support. One simple explanation is that these libraries are optimized for execution on GPUs, and higher-order convolutions require prohibitively large volumes of data with respect to the 16 GB ceiling of today's most advanced GPU hardware. These limitations are clear in previous studies, which either (a) limit the network size~\cite{tran2015learning}, (b) down-sample images to lower resolution~\cite{ji20133d}, or (c) include 3D primitives for only a subset of network layers~\cite{lee2015recursive}.

There are many potential options for circumventing the issue of ConvNet memory usage. The first is to split the network across multiple GPUs, which requires the careful coordination of activation and gradient flow~\cite{dean2012large}. Even in the case of the most successful distributed frameworks for ConvNets~\cite{abadi2016tensorflow}, GPU memory management is largely unresolved. The TensorFlow authors propose two partial solutions warranting further investigation: (a) re-computing versus storing large tensors; and (b) transferring long-lived tensors from GPU to host CPU memory. Instead, we propose an alternative to horizontal scalability for overcoming GPU memory constraints -- a fast implementation of $N$-dimension convolution optimized for multicore CPU systems, which have access to practically unbounded memory on a single node.

\section{Prior Art}
Algorithms for fast convolution have existed in signal processing literature since the 1980s~\cite{winograd1980arithmetic}. The general recipe is to transform both data and kernel into a new space, where expensive sliding window-style convolutions reduce to cheaper element-wise products. The first examples of this approach in ConvNet literature involved Fast Fourier Transforms (FFTs) exploiting the convolution theorem~\cite{mathieu2013fast,vasilache2014fast}. More recently, Lavin and Gray have pioneered the use of the more general class of Winograd-style algorithms~\cite{lavin2016fast,winograd1980arithmetic}. Their implementation and its cuDNN derivatives~\cite{chetlur2014cudnn} have produced state-of-the-art GPU performance on deep networks of small kernels. In this Section we provide a brief overview of the theory underlying this approach, focusing on the aspects that are important for exploiting the architecture of multicore CPUs.

\subsection{Fast Vector Convolution}
\label{ss:vector}
Consider the 1-dimension convolution $\mathbf{s} = \mathbf{g} * \mathbf{d}$, where the kernel and data vectors are of length $G$ and $D$. This problem can be rephrased as one of polynomial multiplication by introducing the associated polynomials $d(x)$, $g(x)$ and $s(x) = g(x)d(x)$, where the coefficients $s_i = \sum_{k}g_{i - k}d_k$ of $x^{i}$ are the solution to the desired convolution. This computation can be distributed across a set of efficient local computations by considering the Chinese Remainder Theorem (CRT)~\cite{ding1996chinese}, as summarized in Theorem 1. By observing that $s(x) = \left[g(x)d(x)\right] \,\mathrm{mod}\, m(x)$ for any polynomial $m(x)$ of sufficiently high degree, we can exploit Theorem 1 to efficiently calculate $s(x)$, as shown in Algorithm 1, which can be conveniently rephrased in terms matrix-vector products:
\begin{equation}
\label{eq:winograd_1d}
\mathbf{s} = \mathbf{A} \left[ \,(\mathbf{Cg}) \odot (\mathbf{Bd}) \, \right],
\end{equation}
where $\mathbf{C}$, $\mathbf{B}$ and $\mathbf{A}$ are introduced as the kernel, data and inverse transforms respectively. With respect to Algorithm 1, Step (1) is implemented by the kernel and data transforms, Step (2) by their transformed element-wise product and Step (3) by the final inverse transform.
\begin{theorem}[CRT for Polynomials]
	 \ \\Let $m(x)=\Pi_{k=1}^{r} m^{(k)}(x)$, where $m^{(k)}(x)$ are pairwise coprime. If $b^{(1)}(x),...,b^{(r)}(x)$ are a set of polynomials then there must exist a unique polynomial $s(x)$ which satisfies the set of congruences:
	\begin{align*}
	s(x) &\equiv b^{(1)}(x)\,\mathrm{mod}\,m^{(1)}(x)  \\
	s(x) &\equiv b^{(2)}(x)\,\mathrm{mod}\,m^{(2)}(x)  \\
	\vdots \\
	s(x) &\equiv b^{(r)}(x)\,\mathrm{mod}\,m^{(r)}(x), 
	\end{align*}
provided the degree of m(x) is not less than that of s(x).
\end{theorem}
\begin{algorithm}[h!]
   \caption{Fast Vector Convolution}
   \label{alg:example}
\begin{algorithmic}
   \STATE {\bfseries Input:} $g(x)$, $d(x)$, $m(x)$
    \FOR{$k=1$ {\bfseries to} $r$}
    	\STATE (1) Compute residual polynomials for $g(x)$ and $d(x)$:
	\begin{align*}
   	g^{(k)}(x) \equiv g(x)\,\mathrm{mod}\,m^{(k)}(x)\\
   	d^{(k)}(x) \equiv d(x)\,\mathrm{mod}\,m^{(k)}(x)
  	 \end{align*}
	 \STATE (2) Compute residual polynomial multiplications:
	\begin{equation*}
   	s^{(k)}(x) = \left[g^{(k)}(x) d^{(k)}(x)\right]\,\mathrm{mod}\,m^{(k)}(x)
   \end{equation*}
   \ENDFOR
     \STATE (3) Reduce partial convolutions to solve $s(x)$:
   \begin{equation*}
   s(x) = \sum_{k=1}^{r} s^{(k)}(x)a^{(k)}(x)
   \end{equation*}
\end{algorithmic}
\end{algorithm}

\subsection{Minimal Winograd Algorithms}
\label{ss:minimal}
In the above formulation (\ref{eq:winograd_1d}), the matrices $\mathbf{C}$ and $\mathbf{B}$ are the remainders of the polynomial divisions $g(x)$ and $d(x)$ by $m^{(k)}(x)$ respectively. The derivation of $\mathbf{A}$ is more involved and is omitted for brevity. Importantly, the only parameter required to synthesize these matrices (in addition to the kernel $g(x)$ and data $d(x)$) is the polynomial $m(x)$.

Traditionally, the selection of $m(x)$ has been subject to many constraints. First, it should be chosen such that the transform matrices contain only degree-1 (scalar) values. Lavin has published code that automates this procedure using the Cook-Toom algorithm to produce transformed kernels and data both of length $D$~\cite{lavin_git}. For an unpadded convolution $\mathbf{s} = \mathbf{g} * \mathbf{d}$ of length $S = D - G + 1$ and ignoring the cost of applying the transforms, this fast algorithm therefore requires $SG/D$ fewer computations to calculate than the standard sliding-window approach. Inappropriate selection of $m(x)$ would yield matrices of polynomials (degree $> 1$) that require considerably more scalar multiplications to compute. 

In reality, the transformations themselves require expensive matrix multiplications that can outweigh the above saving. Accordingly, existing implementations of fast convolution aim to synthesize matrices enriched for ``simple" (e.g. integer) values. There are two motivations for this. First, it improves numeric stability which can have an impact on double-precision convolutions~\cite{lavin2016fast}. More importantly, it supports the hand-crafting of minimal algorithms. These algorithms reduce the cost of applying transform matrices by identifying and eliminating redundant sub-expressions. A famous instance of this approach was documented by Winograd~\cite{winograd1980arithmetic}. Consider the following matrices:
\begin{align*}
\mathbf{A} &= \begin{bmatrix}
       1 & 1 & 1 & 0 \\
       0 & 1 & -1 & -1
     \end{bmatrix} \\
\mathbf{B} &= \begin{bmatrix}
       1 & 0 & -1 & 0 \\
       0 & 1 & 1 & 0 \\
       0 & -1 & 1 & 0 \\
       0 & 1 & 0 & -1
     \end{bmatrix} \quad 
\mathbf{C} = \begin{bmatrix}
       1 & 0 & 0 \\
       \frac{1}{2} & \frac{1}{2} & \frac{1}{2} \\[0.3em]
       \frac{1}{2} & -\frac{1}{2} & \frac{1}{2} \\[0.3em]
       0 & 0 & 1
     \end{bmatrix}.
\end{align*}
By substituting these matrices into (\ref{eq:winograd_1d}) and factoring out redundant computations, we arrive at the following minimal algorithm for vector convolution:
\begin{equation*}
\mathbf{d} * \mathbf{g} = \begin{bmatrix}
       m_1 + m_2 + m_3           \\[0.3em]
       m_2 - m_3 - m_4
     \end{bmatrix},
\end{equation*}
where:
\begin{align*}
m_1 = (d_0 - d_2)g_0, \quad m_2 &= (d_1 + d_2)\frac{g_0 + g_1 + g_2}{2}, \\
m_4 = (d_1 - d_3)g_2,  \quad m_3 &= (d_2 - d_1)\frac{g_0 - g_1 + g_2}{2}.
\end{align*}
This is a so-called $F(S,G)$ algorithm for vector convolution, here for $S = 2$ and $G = 3$. Importantly, this algorithm only works for fixed length kernel and data vectors (here $D = 4$). Generating $F(S, G)$ algorithms for different combinations requires both (a) searching over the space of possible $m(x)$ polynomials as input to Lavin's or similar code~\cite{lavin_git}, and (b) reducing the matrix multiplications to a minimal set of addition, multiplication and shifting operations. To our knowledge there are no automated solutions to either step and thus only a small set of hand-crafted Winograd-style algorithms (e.g. $F(2,3)$, $F(3,4)$ and $F(2,5)$) have been released as fast CPU~\cite{nnpack} or GPU primitives~\cite{chetlur2014cudnn}.

\section{Deep Tensor Convolution}
Below we present an alternative approach to fast convolution that removes the need to hand-craft minimal algorithms. This new approach is better suited to video and volumetric image processing for two main reasons. First, the number of terms involved in a closed-form solution for 3 and higher-dimensional convolutions makes Winograd-style refactoring impractical. Second, by removing numeric simplicity as a constraint we are instead able to synthesize transforms optimized to CPU architectural constraints, e.g. data that are integer multiples of the AVX register width. This is made possible by the relaxed memory constraints of CPUs and allows us to close the previous CPU-GPU performance gap by a full order-of-magnitude.

We first define $N$-dimensional convolution and describe how existing fast algorithms can be extended to this general case. Instead of crafting a minimal algorithm, we show how relaxed memory constraints and efficient sparse linear algebra of CPU systems can be leveraged to amortize transform costs. Later we show how architecture-aware transform synthesis can lead to further acceleration.

\subsection{Convolution in $N$-Dimensions}
Mathematically, the standard convolutional layer used in 2D ConvNets extends trivially to higher-dimensional tensors. Consider a network where for each layer $i$, kernel $j$ and channel $m$
, the kernel weights $\boldsymbol{\mathcal{G}}^{(i,j,m)} = (g\,_{p,q,r})$ and resulting feature map $\boldsymbol{\mathcal{D}}^{(i,j)} = (d\,_{x,y,z})$ are both 3D tensors. This calculation can be expressed element-wise as:
\begin{equation}
\label{eq:3d}
d\,^{(i+1,j)}_{x,y,z} = f\left( b^{(i,j)} + \sum_m \sum_{p, q, r} g\,^{(i,j,m)}_{p,q,r} d\,^{(i,\, j)}_{x+p,\, y+q,\, z+r} \right),
\end{equation}
\noindent{where $ b^{(i,j)}$ is the bias term and $f$ is a ReLU or other non-linear activation function. This extends to higher dimensions by looping over additional subscripts on $g$ and $d$.}

The dimensionality of feature maps is clearly preserved in (\ref{eq:3d}), e.g. a video at the input produces a video at the output. The triple $(p, q, r)$-loop ranges from 0 to the layer-$i$ kernel size to perform sliding-window convolution, and the $m$-loop is a reduction over the previous layer's output channels. This differs from previous studies where the temporal axis is encoded as network channels and flattened after the first layer~\cite{karpathy2014large,simonyan2014two}, producing a single 2D image or class label at the output. These methods have been shown to produce less accurate results on a broad range of video processing tasks when compared to true 3D ConvNets~\cite{tran2015learning}.

It is also evident from (\ref{eq:3d}) why higher-dimensional ConvNets suffer from issues of impractical memory consumption. Each layer of an $N$-dimensional network requires $\boldsymbol{\mathcal{G}}$ and $\boldsymbol{\mathcal{D}}$ to be stored as $N+2$ and $N+1$--dimensional tensors, owing to their operation over multiple kernels and channels. We believe that this multiplicative effect has likely stalled the adoption of the deeper network architectures that dominate image processing tasks, with recent studies instead compromising on network expressiveness to fit within the 16 GB memory constraints of today's top-end GPUs~\cite{ji20133d,lee2015recursive,tran2015learning}.

\subsection{Accelerating Tensor Convolution}
Sidestepping memory constraints by shifting from GPU to CPU hardware is conceptually trivial, as most popular ConvNet frameworks support execution on both CPU and GPU environments. However, the issue preventing the widespread adoption of CPU implementations is not a lack of software support but the large perceived gap between CPU and GPU performance. This is reminiscent of a large ongoing CPU-vs-GPU debate, with various studies claiming that GPUs provide anywhere from 100-to-1000x speed-up across broad problem domains~\cite{lee2010debunking}. A recent review has demonstrated a similar performance gap in the order of 50x across the most popular ConvNet frameworks~\cite{shi2016benchmarking}. Even if distributed GPU solutions like TensorFlow require tensors to be re-computed or swapped between GPU and host CPU memory~\cite{abadi2016tensorflow}, this overhead is easy to justify if the alternative is a 50-fold increase in single-node execution time.

Here we describe how fast algorithms for convolution can be extended to the general case of $N$-dimensional tensors, where the theoretical speed-up is a substantial $(SG/D)^N$. Although recent studies have begun to explore extensions of FFT-based convolution to 3-dimensions~\cite{zlateski2016znni}, to our knowledge there have been no attempts to extend Lavin and Gray's Winograd-style approach~\cite{lavin2016fast}. In order to extend the fast vector algorithm to 1 to $N$-dimensions, we consider the $n$-mode product of a tensor, $\boldsymbol{\mathcal{X}} \in \mathbb{R}^{I_1 \times I_2 \times \dots \times I_N}$, with a matrix, $\mathbf{U} \in \mathbb{R}^{J\times I_n}$, herein denoted as  $\boldsymbol{\mathcal{X}} \times_n \mathbf{U}$~\cite{kolda2009tensor}:
\begin{equation}
\label{eq:tensorprod}
(\boldsymbol{\mathcal{X}} \times_n \mathbf{U})_{i_1, \dots, i_{n-1}, j, i_{n+1}, \dots, i_N} = \sum_{i_n = 1}^{I_n}x_{i_1, \dots, i_N}u_{j, i_n}.
\end{equation}
In our case $\mathbf{U}$ is sparse and $\boldsymbol{\mathcal{X}}$ is dense, so we implement (\ref{eq:tensorprod}) such that $\mathbf{U}$ is traversed in the outermost two loops. We also introduce the following notation for brevity:
\begin{equation*}
\boldsymbol{\mathcal{X}} \times_{n=1}^{N} \mathbf{U}_n = \boldsymbol{\mathcal{X}} \times_{1} \mathbf{U}_1 \times_{2}  \dots \times_{N} \mathbf{U}_N. 
\end{equation*}

The fast algorithm for tensor convolution applies the transforms $\mathbf{C}_n$, $\mathbf{B}_n$ and $\mathbf{A}_n$ separately to each dimension $n$ of the kernel and data tensors, $\boldsymbol{\mathcal{G}}$ and $\boldsymbol{\mathcal{D}}$:
\begin{equation}
\label{eq:winond}
\boldsymbol{\mathcal{S}} =\left[\,( \boldsymbol{\mathcal{G}}\times_{n=1}^{N} \mathbf{C}_{n} )\odot ( \boldsymbol{\mathcal{D}}\times_{n=1}^{N} \mathbf{B}_{n} ) \,\right]\times_{n=1}^{N} \mathbf{A}_{n}.
\end{equation}
It is straightforward to show that (\ref{eq:winograd_1d}) is a special case of (\ref{eq:winond}) by considering the following equivalence:
\begin{equation*}
\label{eq:equiv}
\boldsymbol{\mathcal{Y}} = \boldsymbol{\mathcal{X}} \times_n \mathbf{U} \Leftrightarrow \mathbf{Y}_{(n)} = \mathbf{UX}_{(n)},
\end{equation*}
where the matrix $\mathbf{X}_{(n)}$ is the mode-$n$ major unfolding of tensor $\boldsymbol{\mathcal{X}}$~\cite{kolda2009tensor}. In the 1-dimensional case, $\mathbf{X}_{(1)}$ is simply $\mathbf{x}$ and thus $\boldsymbol{\mathcal{X}} \times_1 \mathbf{U} = \mathbf{Ux}$. Likewise in 2D, as $\boldsymbol{\mathcal{X}}\times_1 \mathbf{U} = \mathbf{UX}$ and $\boldsymbol{\mathcal{X}}\times_2 \mathbf{U} = \mathbf{UX}^\top$ then (\ref{eq:winond}) reduces to the case reported by~\cite{lavin2016fast}:
\begin{equation*}
\mathbf{S} = \mathbf{A} \left[ \,(\mathbf{CGC^\top}) \odot (\mathbf{BDB^\top}) \, \right]\mathbf{A^\top}.
\end{equation*}

\subsection{Amortizing Transform Costs}
Manually reducing transform costs via Winograd-style minimal algorithms is important for 2-dimensional GPU implementations. However, this is less important for a CPU implementation of higher-dimensional convolution. The reasons are two-fold: (a) the matrix multiplication cost can be amortized across a larger number of kernels and channels due to relaxed memory constraints; and (b) CPUs are able to directly leverage the sparse structure of these matrices for further acceleration. Although efficient sparse linear algebra is possible on GPUs, this typically involves reshuffling sparse matrices into a dense representation (e.g. COO, CSR or ELLPACK~\cite{grewe2011automatically}) and introduces unnecessary computational overhead.

As a simple example, consider Winograd's minimal F(2,3) algorithm presented in Section~\ref{ss:minimal}. Computing the output $\mathbf{s}$ of length $S=2$ requires a total of 6 multiplications -- 4 between the data and kernel, and 2 by a constant factor of 0.5. The 4 additions are ignored as modern CPUs can compute fused multiply-accumulate operations in a single cycle. By contrast, computing $\mathbf{s}$ explicitly by equation (\ref{eq:winograd_1d}) requires 28 multiplications -- 4 for the element-wise product, 16 for the data transform and 8 for the inverse transform (assuming transformed kernels are cached at training time). Even leveraging sparsity in the transform matrices requires 19 multiplications, which is more than triple that required for Winograd's minimal algorithm.

The game changes when one considers these approaches in the context of a ConvNet layer with multiple channels and kernels. Without loss of generality, assume the numbers of kernels and channels are both equal to $M$. As the inverse transform can be applied once over the reduced output and the data transform once across all kernels, the required number of multiplications is just $4M^2 + 24M$ (versus $6M^2$ for Winograd). This can be reduced further to $4M^2 + 15M$ by exploiting the sparsity of $\mathbf{A}$ and $\mathbf{B}$. 

\begin{figure}[t]
\begin{center}
\centerline{\includegraphics[width=0.83\columnwidth]{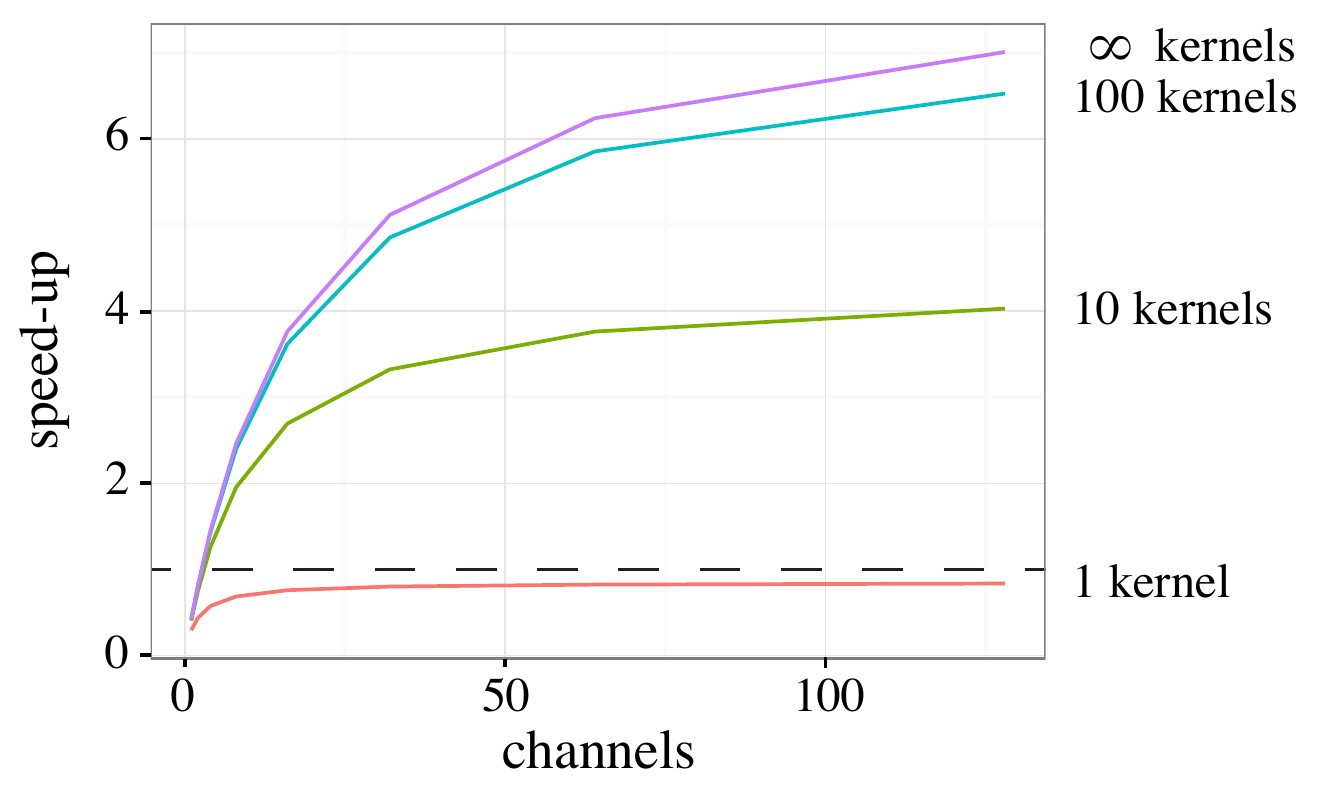}}
\caption{Reduction in computations achieved by fast tensor convolution (forward pass) for a C3D kernel ($3\times3\times3$) as a function of number of layer channels and kernels. Dashed line indicates direct convolution baseline.}
\label{fig:speedup}
\end{center}
\vskip -0.3in
\end{figure} 

Although it is also possible to restructure Winograd's algorithm to exploit the size of the network, for larger networks the $4M^2$ multiplications required by the element-wise product quickly renders the linear transform cost negligible. It is also impractical to construct similar minimal algorithms in higher dimensions. Consider the C3D network of $3\times 3\times 3$ kernels that has yielded state-of-the-art performance across many video processing benchmarks~\cite{tran2015learning}. As an example, we synthesize the following transform matrices such that convolution reduces to a $6\times 6\times 6$ element-wise product:
\begin{align*}
\mathbf{A} &= \begin{bmatrix}
       1  &1  &1   & 1 &     1 &   0\\
       0  &1 & -1 & \frac{1}{3} &  -\frac{1}{3}  & 0\\
       0 & 1&  1  & \frac{1}{9}   & \frac{1}{9} &  0\\
       0  &1 & -1 & \frac{1}{27}&  -\frac{1}{27} & 1
            \end{bmatrix}\\
\mathbf{B} &= \begin{bmatrix}
      \frac{1}{9} &  0   & -\frac{10}{9}&    0 &   1& 0\\
       0  & -\frac{1}{9} & -\frac{1}{9}    & 1   & 1 & 0\\
       0   &\frac{1}{9}   &-\frac{1}{9} &   -1  &  1 & 0\\
       0 &  -\frac{1}{3}  & -1   &  \frac{1}{3} &  1  &0\\
       0&   \frac{1}{3}   & -1   & -\frac{1}{3}  & 1&  0\\
       0  & \frac{1}{9} &    0  &  -\frac{10}{9} & 0  &1
            \end{bmatrix}\\
\mathbf{C} &= \begin{bmatrix}
       9 & \frac{9}{16} & \frac{9}{16} & -\frac{81}{16} & -\frac{81}{16} & 0\\
       0 & \frac{9}{16} & -\frac{9}{16} & -\frac{27}{16} & \frac{27}{16} & 0\\
       0 & \frac{9}{16} & \frac{9}{16} & -\frac{9}{16} & -\frac{9}{16} & 1
            \end{bmatrix}^\top.
\end{align*}
The theoretical ceiling on speed-up obtainable using these matrices is 8-fold, ignoring the cost of the matrix-tensor products required when applying (\ref{eq:winond}). Figure~\ref{fig:speedup} demonstrates the actual reduction in computations as a function of kernels and channels. For a network of just 100 kernels and 100 channels, it is possible to obtain greater than 6-fold acceleration with respect to direct sliding-window convolution. This is triple the performance margin that could be gained if the network was constrained to 10 kernels and channels due to a lower memory ceiling. 

We can further improve this performance margin by exploiting the sparsity of the matrices themselves, as it is comparatively straightforward to implement efficient sparse linear algebra for CPUs. One might worry that the transform matrix sparsity is inversely proportional to the degree of $m(x)$. However, this simply suggests that our fast algorithm is best suited for networks of small kernels, which is fortunately well-aligned with recent trends in deep ConvNet architecture~\cite{he2016deep,simonyan2014very,szegedy2015going}. Sparsity and numerical precision also decrease as a function of $D$. In practice, the data matrix $\textbf{D}$ is not the full feature map (e.g. an ImageNet image) but rather one of many small, overlapping input tiles (each of size $D\times D$, stepping by $S$ along both axes) whose $S\times S$ outputs are stitched together to form the final convolution result. In Section~\ref{ss:avxwino} we discuss how the fully-automated nature of our implementation can leverage this property for further performance improvement.

\section{Optimizing for CPU Architecture}
There are a myriad of algorithmic tricks that can be applied to reduce the number of computations required for convolution. Consider the special case where our transforms are the discrete Fourier transform (DFT) and inverse DFT matrices. As the Fourier transform of a real-valued signal has Hermitian symmetry, the number of unique terms in the element-wise product can be reduced~\cite{mathieu2013fast}. More generally, one could also apply the Strassen algorithm to reduce the number of steps required for matrix multiplication~\cite{cong2014minimizing}. 

In practice, the merit of any of these approaches depends intimately on whether they can be implemented to effectively leverage hardware. Consider the 50-to-1 performance ratio observed between existing GPU and CPU implementations~\cite{shi2016benchmarking}. For the devices used in this study (Titan X versus Xeon E7-8890), the ratio of theoretical throughput is actually less than to 5-to-1. This seems to suggest that current CPU performance limitations are largely issues of software rather than hardware. 

Although some previous studies have discussed CPU-specific performance optimizations for neural networks~\cite{vanhoucke2011improving}, these guidelines have not necessarily translated to optimal implementations. For example, the Eigen 3.2 linear algebra library (used until recently by TensorFlow) does not provide native support for AVX (vectorized) instructions, introducing a tight bottleneck on theoretical throughput. Looking beyond a single core, a recent review demonstrates poor multicore scalability across all major ConvNet frameworks~\cite{shi2016benchmarking}. Solving these two issues alone has the potential to close the CPU-GPU gap by a full order-of-magnitude, and this improvement is multiplicative with the algorithmic savings described earlier.

\subsection{Single-Core Utilization}
Although our fast algorithm requires theoretically fewer computations to execute than na\"{\i}ve convolution (e.g. 8-fold for C3D kernels), it is considerably more difficult to implement with high CPU utilization. 
Consider the element-wise product $\boldsymbol{\mathcal{G}}^\prime \odot \boldsymbol{\mathcal{D}}^\prime$, summed for each channel $m = 1\dots, M$ to produce the $N$-dimensional tensor $\boldsymbol{\mathcal{S}}^\prime$. We can compute the ratio of computations, i.e. 1 multiply and 1 accumulate operation per $(g,\,d)$-pair,  to the volume of memory loaded:
\begin{equation*}
\frac{\mathrm{computations}}{\mathrm{memory\,accesses}} = \frac{2D^NM}{2D^NM} = 1.
\end{equation*}
Little's Law shows this is problematic for effective CPU utilization, as convolution expressed in this form is bottlenecked by memory bandwidth~\cite{little1961proof}. To solve this problem, recall that $\boldsymbol{\mathcal{D}}$ is one of many small, overlapping tiles that span the full-size feature map. Considering $T$ of these tiles, we introduce the following matrices:
\begin{equation}
\label{eq:practical}
\hat{\mathbf{S}}^{(i)} = \hat{\mathbf{D}}^{(i)} \times \hat{\mathbf{G}}^{(i)},
\end{equation}
where $\hat{\mathbf{D}}^{(i)} \in \mathbb{R}^{T\times M}$ (tiles-by-channels) and $\hat{\mathbf{G}}^{(i)} \in \mathbb{R}^{M\times K}$ (channels-by-kernels). Each matrix $i \in 1,\dots, D^N$ captures a single $(x, y)$ coordinate in the earlier $\boldsymbol{\mathcal{G}}^\prime \odot \boldsymbol{\mathcal{D}}^\prime$ element-wise product, which is fused with the channel-wise reduction into end-to-end matrix multiplications:
\begin{equation*}
\frac{\mathrm{computations}}{\mathrm{memory\,accesses}} = \frac{2D^NMTK}{D^N(MT + MK)} = \frac{2\,TK}{T+K}.
\end{equation*}
\begin{algorithm}[t]
   \caption{$N$-Dimensional Convolution with SIMD}
   \label{alg:efficient}
\begin{algorithmic}
    \FOR{$i=1$ {\bfseries by} $W$ {\bfseries to} $D^N$}
    	 \FOR{$m=1$ {\bfseries to} $M$}\vskip 0.05in
	 	\STATE$\texttt{FMA}\left(\hat{\mathbf{s}}_{\,t,\, k}^{(i\,:\,i+W)},\, \hat{\mathbf{d}}_{\,t,\, m}^{(i\,:\,i+W)},\, \hat{\mathbf{g}}_{\,m,\, k}^{(i\,:\,i+W)} \right)$
	 \ENDFOR
    \ENDFOR
\end{algorithmic}
\end{algorithm}
As $T$ can be any number of the small $D^N$ input tiles, we can select $T = K$ to demonstrate a compute-to-memory ratio that grows linearly in the number of kernels. 

\begin{figure}[t]
\begin{center}
\centerline{\includegraphics[width=1.0\columnwidth]{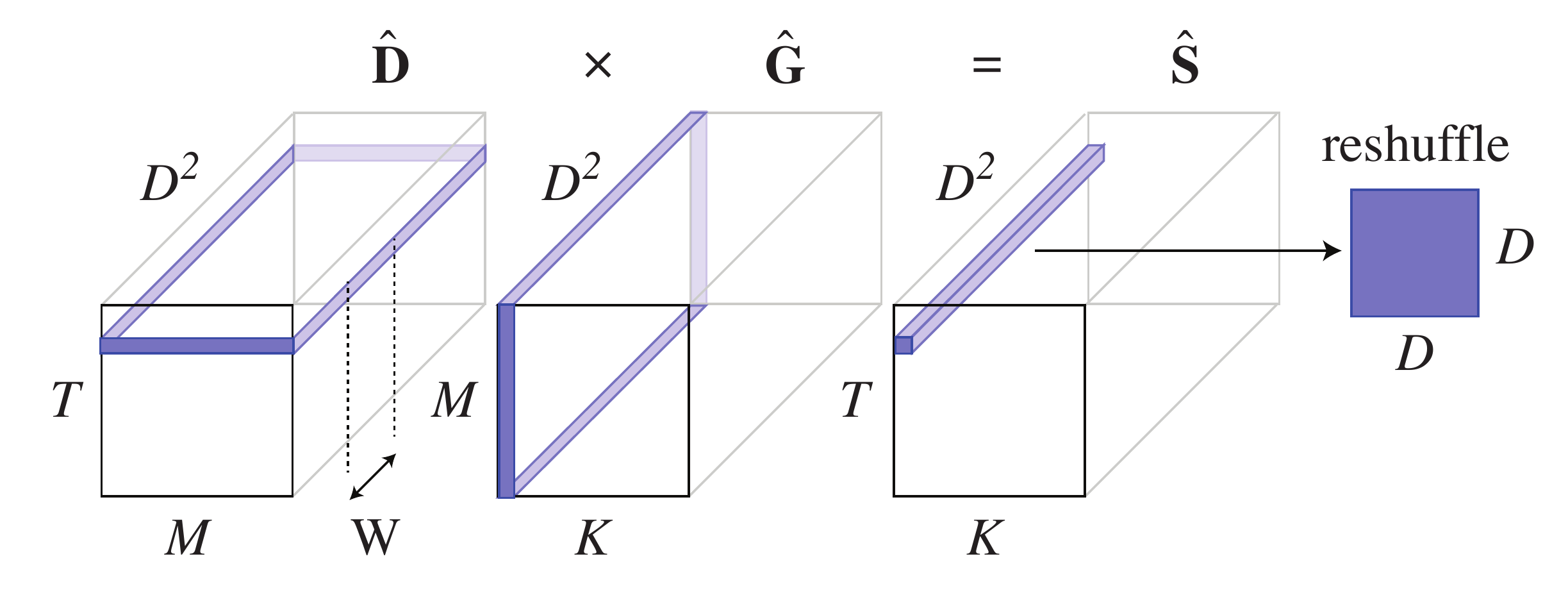}}
\caption{Illustration of Algorithm~\ref{alg:efficient} using 2-dimensional ConvNets as an example. Both the element-wise product $\mathbf{G}^\prime \odot \mathbf{D}^\prime$ and reduction down $M$ channels are captured within matrix multiplication. Multiple elements in $\hat{\mathbf{s}}_{\,t,\, k}$ can be calculated simultaneously by filling AVX registers into-the-page. This technique generalizes trivially to $N$-dimensions by substituting $D^2$ for $D^N$.}
\label{fig:avx}
\end{center}
\vskip -0.3in
\end{figure}

The fast convolutional form in (\ref{eq:practical}) is also well-suited to a number of other practical CPU performance optimizations~\cite{vanhoucke2011improving}. Foremost among these is the effective use of AVX (vectorized) and FMA (fused multiply-accumulate) floating-point SIMD operations. Consider the function \texttt{FMA}($\mathbf{x},\, \mathbf{y},\, \mathbf{z}$), which calculates the sum of vector $\mathbf{x}$ with the element-wise product $\mathbf{y} \odot \mathbf{z}$ and stores the result in $\mathbf{x}$, all in a single CPU cycle. This function can be leveraged for an efficient practical implementation of (\ref{eq:practical}), as presented in Algorithm~\ref{alg:efficient} for a single tile-kernel pair $s_{\,t,\, k}^{(i)} \in \hat{\mathbf{S}}^{(i)}$ and an AVX vector of width $W$. An illustration of the 2-dimensional case is provided in Figure~\ref{fig:avx}. On our Xeon CPU with 256-bit AVX registers and two dedicated FMA units, this optimization alone can yield a 32-fold speed-up over na\"{\i}ve implementations. This margin is expected to double  with the introduction of 512-bit AVX registers for Intel Skylake and Xeon Phi.

We benchmarked the performance of our fast convolution algorithm on a 1.44 TFLOP/s Xeon E7-8890 CPU and observe that it executes at $\sim$70\% maximum utilization. This includes all steps from input to output, including all necessary data reshuffling. As a point of comparison, Intel's own MKL convolutional primitive runs at just $20\%$ (excluding reshuffling) on the same processor. The Eigen 3.2. linear algebra library is lower utilization still, capped at just $3.5\%$ due to a lack of AVX and FMA support. Both of these libraries have been widely used by popular ConvNet frameworks including Caffe, CNTK, TensorFlow and Torch.

\subsection{AVX-Aware Transform Synthesis}
\label{ss:avxwino}
The fully automated nature of our transform generation allows for the synthesis of transform matrices that optimize for CPU architectural constraints. From Figure~\ref{fig:avx}, it is clear the full utilization can only be achieved if $D^N$ is an integer multiple of the AVX vector width $W$. This is an important optimization, as data volumes are constantly small (invariant of numbers of channels and kernels) and thus there is little opportunity to amortize padding overhead. 

Table~\ref{tab:transforms} summarizes statistics for example transforms that we have generated for square 2 and 3-dimensional kernels, enumerated automatically using~\cite{lavin_git}. In each case, we generate transforms for the smallest possible $\mathbf{D}\in\mathbb{R}^{D\times D}$ such that $SG/D > 1$ and $D^2 \,\mathrm{mod}\, W = 0$.  The matrices are provided in the Supplementary Materials.

\begin{table}[t]
\centering
\caption{Size, transform sparsity and algorithmic speed-up statistics for example transforms matrices. Associated matrices are provided in the Supplementary Materials.\vspace{1em}} 
\label{tab:transforms}
\begin{tabular}{c|ccc|ccc|cc}
    & \multicolumn{3}{c|}{\textbf{size}} & \multicolumn{3}{c|}{\textbf{sparsity}} & \multicolumn{2}{c}{\textbf{speed-up}} \\
    & $D$          & $G$         & $S$         & $\mathbf{A}$           & $\mathbf{B}$           & $\mathbf{D}$          & 2D                & 3D                \\ \hline
(a) & 4          & 2         & 3         & 0.33        & 0.50        & 0.25       & 2.25              & 3.38              \\
(b) & 4          & 3         & 2         & 0.25        & 0.50        & 0.33       & 2.25              & 3.38              \\
(c) & 8          & 4         & 5         & 0.20        & 0.31        & 0.19       & 6.25              & 15.63             \\
(d) & 8          & 5         & 4         & 0.19        & 0.31        & 0.20       & 6.25              & 15.63             \\
(e) & 8          & 6         & 3         & 0.17        & 0.31        & 0.21       & 5.06              & 11.39            
\end{tabular}
\end{table}

\subsection{Multicore Scalability}
Single-core utilization is just one dimension of performance optimization. Many modern systems contain both multiple CPU chips, with shared access to host RAM; and multiple cores per chip, with shared access to faster L3 cache. We adopt a relatively simple parallelization scheme where threads simultaneously operate on different subsets of $T$ input tiles. To avoid memory contention and other concurrency issues we adopt the Cilk Plus work-stealing scheduler supported by GCC 4.8~\cite{blumofe1996cilk,robison2013composable}, simply applying its fork-join primitive to all for-loops with no iteration dependencies. The number of tiles $T$ per thread is empirically tuned to simultaneously maximize L3 cache utilization ($T$ cannot be too large) and compute-to-memory ratio ($T$ cannot be too small).

\begin{figure}[t]
\begin{center}
\centerline{\includegraphics[width=0.92\columnwidth]{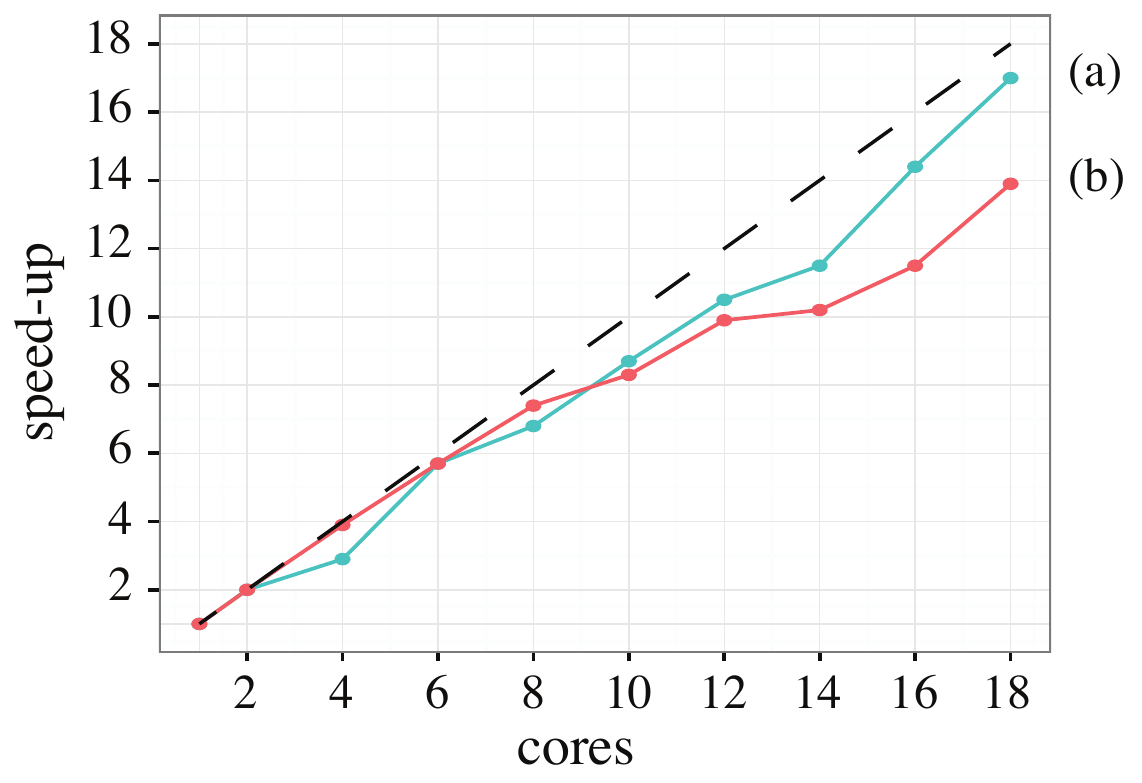}}
\caption{Multicore scalability of our cache-aware and Cilk-optimized implementations of (a) fast convolution, and (b) na\"{\i}ve convolution. Dashed line indicates theoretical scalability limit with respect to a single-core implementation. Executed on 18-core Intel Xeon E7-8890 processor with 45 MB L3-cache.}
\label{fig:multicore}
\end{center}
\vskip -0.3in
\end{figure}

We observe that even this simple parallelization scheme yields near-optimal linear scalability. In Figure~\ref{fig:multicore} we present ConvNet throughput as a function of processor cores for both (a) our fast algorithm, and (b) our own multicore implementation of na\"{\i}ve convolution (which is comparatively simple to implement). Scalability is measured across a single convolution layer for a $1024\times 1024$ image with kernels of size $4\times 4$. To avoid NUMA issues relating to expensive inter-chip communication, we spawn independent instances for each CPU in our 4-socket shared-memory server such that all 18 threads in Figure~\ref{fig:multicore} are bound to a single chip. When using all 18 cores of our Intel Xeon E7-8890 CPU the scalability of (a) is 95\% theoretically optimal. As a point of comparison, a recent review examined the scalability of popular ConvNet frameworks Caffe, CNTK, TensorFlow and Torch on a similar 16-core Xeon E5-2630 CPU~\cite{shi2016benchmarking}. They reported multicore scalability ranging from $16\%$ (Caffe) to $42\%$ (TensorFlow), which is equivalent to a 2.3 to 5.9-fold improvement with our implementation.

\subsection{Performance Benchmarking}
The most popular ConvNet benchmarks focus exclusively on GPU performance~\cite{chintala2015convnet}. The only study we could find presenting thorough CPU benchmarking is that of Shi \emph{et al.}, comparing the throughput of Caffe, CNTK, Tensorflow and Torch for the AlexNet and ResNet architectures~\cite{shi2016benchmarking}. Although this is a useful study for ball-parking our multicore scalability, it is difficult to extrapolate fair comparisons to our overall system throughput for many reasons. Foremost is that the authors do not select CPU-optimized implementations. They adopt an earlier version of TensorFlow that uses the Eigen 3.2 library (no AVX/FMA support), and otherwise use the default framework-specific implementations of convolution rather than linking to optimized packages such as Intel MKL.

We benchmark 2D ConvNet performance against two popular frameworks: TensorFlow, using the newer Eigen 3.3 library (with AVX support); and Caffe, compiled to use Intel's optimized MKL library. We consider the propagation time of a $224\times224$ ImageNet image through three  convolution layers to capture any necessary inter-layer reshuffling. We choose this simple architecture over a named network because we are not interested in comparing execution times of pooling, fully-connected or other layers. We also select an obscure kernel size ($4\times4$) for which there have been no Winograd-style fast algorithms published, in order to demonstrate the generality of our implementation to arbitrary kernels. Each layer contains a modest 32 channels and 32 kernels for spreading the cost associated with applying transform matrices. Results presented are the fastest across batch sizes of 1, 20 and 200. An important innovation of our approach is that it is batch size-agnostic, making it suitable for single-image autoregressive models common in generative modelling and deep reinforcement learning.
\begin{figure}[t]
\begin{center}
\centerline{\includegraphics[width=0.85\columnwidth]{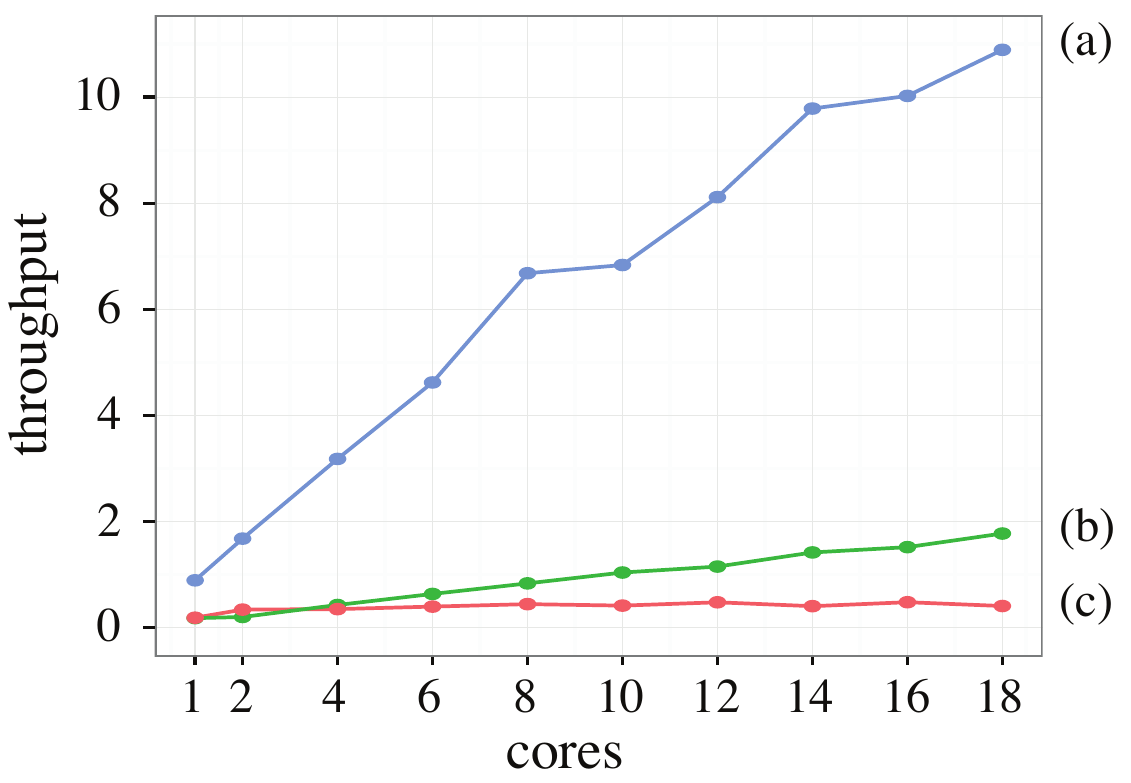}}
\caption{Measured throughput (megavoxels per second) of (a) our fast 2D convolution implementation (as a special case of our $N$-dimensional algorithm), (b) TensorFlow, using the latest Eigen 3.3, and (c) Caffe, using Intel MKL. Throughput is calculated by propagating $224\times 224$ images through 3 convolutional layers.}
\label{fig:tensorflow}
\end{center}
\vskip -0.3in
\end{figure} 

Our performance benchmarks are presented in Figure~\ref{fig:tensorflow}. The single-core throughput of (a) our fast algorithm is 0.89 MVox/s, compared to (b) 0.18 for TensorFlow and (c) 0.19 for Caffe. Increasing cores from 1 to 18, our throughput improves to 10.9 MVox/s compared to 1.77 for TensorFlow and 0.41 for Caffe. This is equivalent to an approximate 5 to 25-fold improvement in overall performance. In terms of multicore scalability, this is (a) 68\% versus (b) 55\% and (c) 12\%. We note that our performance here is lower than the $95\%$ presented in Figure~\ref{fig:multicore} for a larger input size (i.e. $T$ is much larger, yielding a better compute-to-memory ratio), and that the scalability for TensorFlow and Caffe are both similar to those reported in~\cite{shi2016benchmarking}.

\section{Discussion}
Motivated by the recent success of 3-dimensional ConvNets in video and volumetric image processing~\cite{lee2015recursive,tran2015learning}, we have proposed a transition to CPU hardware to overcome the memory constraints limiting the size and expressivity of these networks. Key to this transition is overcoming the impractical performance gap between existing CPU and GPU implementations. To achieve this, we extended previous algorithms of fast convolution to the $N$-dimensional case, yielding an order-of-magnitude reduction in computations for popular networks such as C3D. Importantly, our implementation diverges from previous studies that focus on the hand-crafting of minimal Winograd-style algorithms. We instead exploit the relaxed memory constraints, efficient sparse access and other architectural considerations of CPU hardware to overcome the cost of applying transform matrices.

The obvious alternative to our approach is to overcome memory constraints by splitting large networks across multiple GPU devices. Distributed frameworks such as TensorFlow are valuable for a broad class of machine learning problems, e.g. many of the data mining tasks faced by large organizations where the data itself is often sharded across different machines. However, it is important to recognize that the horizontal scalability paradigm is not a one-size-fits-all solution. Consider the increasing demand for real-time CPU solutions to image and video processing, particularly on mobile devices. Moving forward, we expect that intensive ConvNet-driven tasks such as video classification and de-noising will continue to migrate from the realm of academic research to practical realization~\cite{shi2016real}. Efficient CPU implementations of ConvNets and other deep learning algorithms will play a fundamental role in this transition.

At the opposite end of the spectrum, some ``big data" problems in the image processing domain are, counterintuitively, too big to be solved in a distributed setting. Consider the emerging field of high-throughput connectomics~\cite{meirovitch2016multi}. Multi-beam electron microscopes image cross-sectional slices of neural tissue at nanometer-resolution, which are then segmented by ConvNets to reconstruct the 3-dimensional morphology and interconnectivity of individual neurons~\cite{ronneberger2015u}. The major issue here is simply one of scale -- a seemingly modest cubic millimeter volume of neural tissue takes several months to image at the TB/hr pace of modern electron microscopes, which exceeds maximum data transfer rates. To avoid introducing communication bottlenecks to the connectomics pipeline, it is necessary that segmentation can execute in real-time on a server physically co-located in the same room as the microscope~\cite{lichtman2014big,matveev2016}. Shared-memory CPU systems can support hundreds of cores and terabytes of memory in a single server, and it is critical that systems be implemented to exploit these valuable resources.

Treating 2D ConvNets as a special case of tensor convolution, our implementation yields 5 to 25-fold improved throughput compared to previous state-of-the-art on CPU. This is an important step toward bridging the performance gap between CPU and GPU hardware and is particularly important in the context of emerging hardware trends, e.g. Intel announcing that future generations of CPUs will contain dedicated deep learning accelerators. More importantly, we believe that removing constraints on 3D ConvNet size will herald new opportunities in the machine learning community; particularly in the context of generative models~\cite{denton2015deep,goodfellow2014generative}, where rich temporal correlations are currently ignored when learning latent manifolds~\cite{ledig2016photo}.

\section*{Acknowledgements} 
Support is gratefully acknowledged from the National Science Foundation (NSF) under grants IIS-1447786 and CCF-1563880, and the Intelligence Advanced Research Projects Activity (IARPA) under grant 138076-5093555.

\bibliography{refs}

\begin{thebibliography}{46}
\providecommand{\natexlab}[1]{#1}
\providecommand{\url}[1]{\texttt{#1}}
\expandafter\ifx\csname urlstyle\endcsname\relax
  \providecommand{\doi}[1]{doi: #1}\else
  \providecommand{\doi}{doi: \begingroup \urlstyle{rm}\Url}\fi

\bibitem[Abadi et~al.(2016)Abadi, Agarwal, Barham, Brevdo, Chen, Citro,
  Corrado, Davis, Dean, Devin, et~al.]{abadi2016tensorflow}
Abadi, Mart{\i}n, Agarwal, Ashish, Barham, Paul, Brevdo, Eugene, Chen, Zhifeng,
  Citro, Craig, Corrado, Greg~S, Davis, Andy, Dean, Jeffrey, Devin, Matthieu,
  et~al.
\newblock Tensorflow: Large-scale machine learning on heterogeneous distributed
  systems.
\newblock \emph{arXiv preprint arXiv:1603.04467}, 2016.

\bibitem[Blumofe et~al.(1996)Blumofe, Joerg, Kuszmaul, Leiserson, Randall, and
  Zhou]{blumofe1996cilk}
Blumofe, Robert~D, Joerg, Christopher~F, Kuszmaul, Bradley~C, Leiserson,
  Charles~E, Randall, Keith~H, and Zhou, Yuli.
\newblock Cilk: An efficient multithreaded runtime system.
\newblock \emph{Journal of parallel and distributed computing}, 37\penalty0
  (1):\penalty0 55--69, 1996.

\bibitem[Chetlur et~al.(2014)Chetlur, Woolley, Vandermersch, Cohen, Tran,
  Catanzaro, and Shelhamer]{chetlur2014cudnn}
Chetlur, Sharan, Woolley, Cliff, Vandermersch, Philippe, Cohen, Jonathan, Tran,
  John, Catanzaro, Bryan, and Shelhamer, Evan.
\newblock cudnn: Efficient primitives for deep learning.
\newblock \emph{arXiv preprint arXiv:1410.0759}, 2014.

\bibitem[Chintala(2015)]{chintala2015convnet}
Chintala, Soumith.
\newblock Convnet benchmarks.
\newblock github.com/soumith/convnet-benchmarks, 2015.

\bibitem[Cong \& Xiao(2014)Cong and Xiao]{cong2014minimizing}
Cong, Jason and Xiao, Bingjun.
\newblock Minimizing computation in convolutional neural networks.
\newblock In \emph{International Conference on Artificial Neural Networks},
  pp.\  281--290. Springer, 2014.

\bibitem[Couprie et~al.(2013)Couprie, Farabet, Najman, and
  LeCun]{couprie2013indoor}
Couprie, Camille, Farabet, Cl{\'e}ment, Najman, Laurent, and LeCun, Yann.
\newblock Indoor semantic segmentation using depth information.
\newblock \emph{arXiv preprint arXiv:1301.3572}, 2013.

\bibitem[Dean et~al.(2012)Dean, Corrado, Monga, Chen, Devin, Mao, Senior,
  Tucker, Yang, Le, et~al.]{dean2012large}
Dean, Jeffrey, Corrado, Greg, Monga, Rajat, Chen, Kai, Devin, Matthieu, Mao,
  Mark, Senior, Andrew, Tucker, Paul, Yang, Ke, Le, Quoc~V, et~al.
\newblock Large scale distributed deep networks.
\newblock In \emph{Advances in neural information processing systems}, pp.\
  1223--1231, 2012.

\bibitem[Denton et~al.(2015)Denton, Chintala, Fergus, et~al.]{denton2015deep}
Denton, Emily~L, Chintala, Soumith, Fergus, Rob, et~al.
\newblock Deep generative image models using a laplacian pyramid of adversarial
  networks.
\newblock In \emph{Advances in neural information processing systems}, pp.\
  1486--1494, 2015.

\bibitem[Ding et~al.(1996)Ding, Pei, and Salomaa]{ding1996chinese}
Ding, Cunsheng, Pei, Dingyi, and Salomaa, Arto.
\newblock \emph{Chinese remainder theorem: applications in computing, coding,
  cryptography}.
\newblock World Scientific, 1996.

\bibitem[Dukhan(2016)]{nnpack}
Dukhan, M.
\newblock Nnpack.
\newblock \url{https://github.com/Maratyszcza/NNPACK}, 2016.

\bibitem[Glorot \& Bengio(2010)Glorot and Bengio]{glorot2010understanding}
Glorot, Xavier and Bengio, Yoshua.
\newblock Understanding the difficulty of training deep feedforward neural
  networks.
\newblock In \emph{Aistats}, volume~9, pp.\  249--256, 2010.

\bibitem[Goodfellow et~al.(2014)Goodfellow, Pouget-Abadie, Mirza, Xu,
  Warde-Farley, Ozair, Courville, and Bengio]{goodfellow2014generative}
Goodfellow, Ian, Pouget-Abadie, Jean, Mirza, Mehdi, Xu, Bing, Warde-Farley,
  David, Ozair, Sherjil, Courville, Aaron, and Bengio, Yoshua.
\newblock Generative adversarial nets.
\newblock In \emph{Advances in Neural Information Processing Systems}, pp.\
  2672--2680, 2014.

\bibitem[Grewe \& Lokhmotov(2011)Grewe and Lokhmotov]{grewe2011automatically}
Grewe, Dominik and Lokhmotov, Anton.
\newblock Automatically generating and tuning gpu code for sparse matrix-vector
  multiplication from a high-level representation.
\newblock In \emph{Proceedings of the Fourth Workshop on General Purpose
  Processing on Graphics Processing Units}, pp.\ ~12. ACM, 2011.

\bibitem[He et~al.(2016)He, Zhang, Ren, and Sun]{he2016deep}
He, Kaiming, Zhang, Xiangyu, Ren, Shaoqing, and Sun, Jian.
\newblock Deep residual learning for image recognition.
\newblock In \emph{The IEEE Conference on Computer Vision and Pattern
  Recognition (CVPR)}, June 2016.

\bibitem[Ji et~al.(2013)Ji, Xu, Yang, and Yu]{ji20133d}
Ji, Shuiwang, Xu, Wei, Yang, Ming, and Yu, Kai.
\newblock 3d convolutional neural networks for human action recognition.
\newblock \emph{IEEE transactions on pattern analysis and machine
  intelligence}, 35\penalty0 (1):\penalty0 221--231, 2013.

\bibitem[Jia et~al.(2014)Jia, Shelhamer, Donahue, Karayev, Long, Girshick,
  Guadarrama, and Darrell]{jia2014caffe}
Jia, Yangqing, Shelhamer, Evan, Donahue, Jeff, Karayev, Sergey, Long, Jonathan,
  Girshick, Ross, Guadarrama, Sergio, and Darrell, Trevor.
\newblock Caffe: Convolutional architecture for fast feature embedding.
\newblock In \emph{Proceedings of the 22nd ACM international conference on
  Multimedia}, pp.\  675--678. ACM, 2014.

\bibitem[Karpathy et~al.(2014)Karpathy, Toderici, Shetty, Leung, Sukthankar,
  and Fei-Fei]{karpathy2014large}
Karpathy, Andrej, Toderici, George, Shetty, Sanketh, Leung, Thomas, Sukthankar,
  Rahul, and Fei-Fei, Li.
\newblock Large-scale video classification with convolutional neural networks.
\newblock In \emph{Proceedings of the IEEE conference on Computer Vision and
  Pattern Recognition}, pp.\  1725--1732, 2014.

\bibitem[Kolda \& Bader(2009)Kolda and Bader]{kolda2009tensor}
Kolda, Tamara~G and Bader, Brett~W.
\newblock Tensor decompositions and applications.
\newblock \emph{SIAM review}, 51\penalty0 (3):\penalty0 455--500, 2009.

\bibitem[Krizhevsky et~al.(2012)Krizhevsky, Sutskever, and
  Hinton]{krizhevsky2012imagenet}
Krizhevsky, Alex, Sutskever, Ilya, and Hinton, Geoffrey~E.
\newblock Imagenet classification with deep convolutional neural networks.
\newblock In \emph{Advances in neural information processing systems}, pp.\
  1097--1105, 2012.

\bibitem[Lavin(2016)]{lavin_git}
Lavin, A.
\newblock Wincnn.
\newblock \url{https://github.com/andravin/wincnn}, 2016.

\bibitem[Lavin \& Gray(2016)Lavin and Gray]{lavin2016fast}
Lavin, Andrew and Gray, Scott.
\newblock Fast algorithms for convolutional neural networks.
\newblock In \emph{The IEEE Conference on Computer Vision and Pattern
  Recognition (CVPR)}, June 2016.

\bibitem[LeCun et~al.(1989)LeCun, Boser, Denker, Henderson, Howard, Hubbard,
  and Jackel]{lecun1989backpropagation}
LeCun, Yann, Boser, Bernhard, Denker, John~S, Henderson, Donnie, Howard,
  Richard~E, Hubbard, Wayne, and Jackel, Lawrence~D.
\newblock Backpropagation applied to handwritten zip code recognition.
\newblock \emph{Neural computation}, 1\penalty0 (4):\penalty0 541--551, 1989.

\bibitem[LeCun et~al.(2012)LeCun, Bottou, Orr, and
  M{\"u}ller]{lecun2012efficient}
LeCun, Yann~A, Bottou, L{\'e}on, Orr, Genevieve~B, and M{\"u}ller,
  Klaus-Robert.
\newblock Efficient backprop.
\newblock In \emph{Neural networks: Tricks of the trade}, pp.\  9--48.
  Springer, 2012.

\bibitem[Ledig et~al.(2016)Ledig, Theis, Husz{\'a}r, Caballero, Aitken, Tejani,
  Totz, Wang, and Shi]{ledig2016photo}
Ledig, Christian, Theis, Lucas, Husz{\'a}r, Ferenc, Caballero, Jose, Aitken,
  Andrew, Tejani, Alykhan, Totz, Johannes, Wang, Zehan, and Shi, Wenzhe.
\newblock Photo-realistic single image super-resolution using a generative
  adversarial network.
\newblock \emph{arXiv preprint arXiv:1609.04802}, 2016.

\bibitem[Lee et~al.(2015)Lee, Zlateski, Ashwin, and Seung]{lee2015recursive}
Lee, Kisuk, Zlateski, Aleksandar, Ashwin, Vishwanathan, and Seung, H~Sebastian.
\newblock Recursive training of 2d-3d convolutional networks for neuronal
  boundary prediction.
\newblock In \emph{Advances in Neural Information Processing Systems}, pp.\
  3573--3581, 2015.

\bibitem[Lee et~al.(2010)Lee, Kim, Chhugani, Deisher, Kim, Nguyen, Satish,
  Smelyanskiy, Chennupaty, Hammarlund, et~al.]{lee2010debunking}
Lee, Victor~W, Kim, Changkyu, Chhugani, Jatin, Deisher, Michael, Kim, Daehyun,
  Nguyen, Anthony~D, Satish, Nadathur, Smelyanskiy, Mikhail, Chennupaty,
  Srinivas, Hammarlund, Per, et~al.
\newblock Debunking the 100x gpu vs. cpu myth: an evaluation of throughput
  computing on cpu and gpu.
\newblock \emph{ACM SIGARCH Computer Architecture News}, 38\penalty0
  (3):\penalty0 451--460, 2010.

\bibitem[Lichtman et~al.(2014)Lichtman, Pfister, and Shavit]{lichtman2014big}
Lichtman, Jeff~W, Pfister, Hanspeter, and Shavit, Nir.
\newblock The big data challenges of connectomics.
\newblock \emph{Nature neuroscience}, 17\penalty0 (11):\penalty0 1448--1454,
  2014.

\bibitem[Little(1961)]{little1961proof}
Little, John~DC.
\newblock A proof for the queuing formula: L= $\lambda$ w.
\newblock \emph{Operations research}, 9\penalty0 (3):\penalty0 383--387, 1961.

\bibitem[Long et~al.(2015)Long, Shelhamer, and Darrell]{long2015fully}
Long, Jonathan, Shelhamer, Evan, and Darrell, Trevor.
\newblock Fully convolutional networks for semantic segmentation.
\newblock In \emph{Proceedings of the IEEE Conference on Computer Vision and
  Pattern Recognition}, pp.\  3431--3440, 2015.

\bibitem[Mathieu et~al.(2013)Mathieu, Henaff, and LeCun]{mathieu2013fast}
Mathieu, Michael, Henaff, Mikael, and LeCun, Yann.
\newblock Fast training of convolutional networks through {FFTs}.
\newblock \emph{arXiv preprint arXiv:1312.5851}, 2013.

\bibitem[Matveev et~al.(2017)Matveev, Meirovitch, Saribekyan, Jakubiuk, Kaler,
  Odor, Budden, Zlateski, and Shavit]{matveev2016}
Matveev, Alexander, Meirovitch, Yaron, Saribekyan, Hayk, Jakubiuk, Wiktor,
  Kaler, Tim, Odor, Gergely, Budden, David, Zlateski, Aleksandar, and Shavit,
  Nir.
\newblock A multicore path to connectomics-on-demand.
\newblock In \emph{Proceedings of the 22nd ACM SIGPLAN Symposium on Principles
  and Practice of Parallel Programming}. ACM, 2017.

\bibitem[Meirovitch et~al.(2016)Meirovitch, Matveev, Saribekyan, Budden,
  Rolnick, Odor, Jones, Pfister, Lichtman, and Shavit]{meirovitch2016multi}
Meirovitch, Yaron, Matveev, Alexander, Saribekyan, Hayk, Budden, David,
  Rolnick, David, Odor, Gergely, Jones, Seymour Knowles-Barley Thouis~Raymond,
  Pfister, Hanspeter, Lichtman, Jeff~William, and Shavit, Nir.
\newblock A multi-pass approach to large-scale connectomics.
\newblock \emph{arXiv preprint arXiv:1612.02120}, 2016.

\bibitem[Poole et~al.(2016)Poole, Lahiri, Raghu, Sohl-Dickstein, and
  Ganguli]{poole2016exponential}
Poole, Ben, Lahiri, Subhaneil, Raghu, Maithra, Sohl-Dickstein, Jascha, and
  Ganguli, Surya.
\newblock Exponential expressivity in deep neural networks through transient
  chaos.
\newblock \emph{arXiv preprint arXiv:1606.05340}, 2016.

\bibitem[Robison(2013)]{robison2013composable}
Robison, Arch~D.
\newblock Composable parallel patterns with intel cilk plus.
\newblock \emph{Computing in Science and Engineering}, 15\penalty0
  (2):\penalty0 66--71, 2013.

\bibitem[Ronneberger et~al.(2015)Ronneberger, Fischer, and
  Brox]{ronneberger2015u}
Ronneberger, Olaf, Fischer, Philipp, and Brox, Thomas.
\newblock U-net: Convolutional networks for biomedical image segmentation.
\newblock In \emph{International Conference on Medical Image Computing and
  Computer-Assisted Intervention}, pp.\  234--241. Springer, 2015.

\bibitem[Russakovsky et~al.(2015)Russakovsky, Deng, Su, Krause, Satheesh, Ma,
  Huang, Karpathy, Khosla, Bernstein, et~al.]{russakovsky2015imagenet}
Russakovsky, Olga, Deng, Jia, Su, Hao, Krause, Jonathan, Satheesh, Sanjeev, Ma,
  Sean, Huang, Zhiheng, Karpathy, Andrej, Khosla, Aditya, Bernstein, Michael,
  et~al.
\newblock Imagenet large scale visual recognition challenge.
\newblock \emph{International Journal of Computer Vision}, 115\penalty0
  (3):\penalty0 211--252, 2015.

\bibitem[Shi et~al.(2016{\natexlab{a}})Shi, Wang, Xu, and
  Chu]{shi2016benchmarking}
Shi, Shaohuai, Wang, Qiang, Xu, Pengfei, and Chu, Xiaowen.
\newblock Benchmarking state-of-the-art deep learning software tools.
\newblock \emph{arXiv preprint arXiv:1608.07249}, 2016{\natexlab{a}}.

\bibitem[Shi et~al.(2016{\natexlab{b}})Shi, Caballero, Husz{\'a}r, Totz,
  Aitken, Bishop, Rueckert, and Wang]{shi2016real}
Shi, Wenzhe, Caballero, Jose, Husz{\'a}r, Ferenc, Totz, Johannes, Aitken,
  Andrew~P, Bishop, Rob, Rueckert, Daniel, and Wang, Zehan.
\newblock Real-time single image and video super-resolution using an efficient
  sub-pixel convolutional neural network.
\newblock In \emph{Proceedings of the IEEE Conference on Computer Vision and
  Pattern Recognition}, pp.\  1874--1883, 2016{\natexlab{b}}.

\bibitem[Simonyan \& Zisserman(2014{\natexlab{a}})Simonyan and
  Zisserman]{simonyan2014two}
Simonyan, Karen and Zisserman, Andrew.
\newblock Two-stream convolutional networks for action recognition in videos.
\newblock In \emph{Advances in Neural Information Processing Systems}, pp.\
  568--576, 2014{\natexlab{a}}.

\bibitem[Simonyan \& Zisserman(2014{\natexlab{b}})Simonyan and
  Zisserman]{simonyan2014very}
Simonyan, Karen and Zisserman, Andrew.
\newblock Very deep convolutional networks for large-scale image recognition.
\newblock \emph{arXiv preprint arXiv:1409.1556}, 2014{\natexlab{b}}.

\bibitem[Szegedy et~al.(2015)Szegedy, Liu, Jia, Sermanet, Reed, Anguelov,
  Erhan, Vanhoucke, and Rabinovich]{szegedy2015going}
Szegedy, Christian, Liu, Wei, Jia, Yangqing, Sermanet, Pierre, Reed, Scott,
  Anguelov, Dragomir, Erhan, Dumitru, Vanhoucke, Vincent, and Rabinovich,
  Andrew.
\newblock Going deeper with convolutions.
\newblock In \emph{Proceedings of the IEEE Conference on Computer Vision and
  Pattern Recognition}, pp.\  1--9, 2015.

\bibitem[Tran et~al.(2015)Tran, Bourdev, Fergus, Torresani, and
  Paluri]{tran2015learning}
Tran, Du, Bourdev, Lubomir, Fergus, Rob, Torresani, Lorenzo, and Paluri,
  Manohar.
\newblock Learning spatiotemporal features with 3d convolutional networks.
\newblock In \emph{2015 IEEE International Conference on Computer Vision
  (ICCV)}, pp.\  4489--4497. IEEE, 2015.

\bibitem[Vanhoucke et~al.(2011)Vanhoucke, Senior, and
  Mao]{vanhoucke2011improving}
Vanhoucke, Vincent, Senior, Andrew, and Mao, Mark~Z.
\newblock Improving the speed of neural networks on {CPU}s.
\newblock 2011.

\bibitem[Vasilache et~al.(2014)Vasilache, Johnson, Mathieu, Chintala, Piantino,
  and LeCun]{vasilache2014fast}
Vasilache, Nicolas, Johnson, Jeff, Mathieu, Michael, Chintala, Soumith,
  Piantino, Serkan, and LeCun, Yann.
\newblock Fast convolutional nets with fbfft: A gpu performance evaluation.
\newblock \emph{arXiv preprint arXiv:1412.7580}, 2014.

\bibitem[Winograd(1980)]{winograd1980arithmetic}
Winograd, Shmuel.
\newblock \emph{Arithmetic complexity of computations}, volume~33.
\newblock Siam, 1980.

\bibitem[Zlateski et~al.(2016)Zlateski, Lee, and Seung]{zlateski2016znni}
Zlateski, Aleksandar, Lee, Kisuk, and Seung, H~Sebastian.
\newblock Znni-maximizing the inference throughput of 3d convolutional networks
  on multi-core cpus and gpus.
\newblock \emph{arXiv preprint arXiv:1606.05688}, 2016.

\end{thebibliography}
\bibliographystyle{icml2017}

\end{document}